\documentclass[letterpaper, 9 pt, conference]{ieeeconf}  
\usepackage{graphicx}
\usepackage{algorithm}
\usepackage{algorithmic}
\usepackage{amsmath}
\usepackage{amssymb}

\IEEEoverridecommandlockouts                              

\title{\LARGE \bf
Query-based Semantic Gaussian Field for Scene Representation in Reinforcement Learning
}

\author{Jiaxu Wang$^{1}$, Ziyi Zhang$^{1}$, Qiang Zhang$^{1, 2,\clubsuit}$, Jingkai Sun$^{1, 2}$, Jiahang Cao$^{1}$, Mingyuan Sun$^{3}$ \\ Junhao He$^{1}$, Hao Cheng$^{1}$, Yi Gu$^{1}$, Renjing Xu$^{1 \dagger}$
\thanks{$^{1}$Hong Kong University of Science and Technology
        {\tt\small \{jwang457, jsun444, jcao248, jhe382, ygu425\}@connect.hkust-gz.edu.cn},
        {\tt\small \{ziyizhang, renjingxu\}@hkust-gz.edu.cn}}
\thanks{$^{2}$Beijing Innovation Center of Humanoid Robotics Co. Ltd.
        {\tt\small Jony.Zhang@xhumanoid.com}}
\thanks{$^{3}$Northeastern University.
        {\tt\small mingyuansun@stumail.neu.edu.cn} $\clubsuit$ is the project leader, $\dagger$ is the corresponding author}
}

\begin{document}

\maketitle
\thispagestyle{empty}
\pagestyle{empty}

\begin{abstract}
Latent scene representation plays a significant role in training reinforcement learning (RL) agents. Recent works incorporate the 3D-aware latent-conditioned NeRF pipeline into scene representation learning to obtain good latent vectors describing the scenes. However, these NeRF-related methods struggle to perceive 3D structural information due to the inefficient dense sampling in volumetric rendering. Moreover, they lack fine-grained semantic information included in their scene representation vectors because they evenly consider free and occupied spaces. Both of them can destroy the performance of downstream RL tasks. To address the above challenges, we propose a novel framework that adopts the efficient 3D Gaussian Splatting (3DGS) to learn 3D scene representation for the first time. In brief, we present the Query-based Generalizable 3DGS to bridge the 3DGS technique and scene representations with more geometrical awareness than those in NeRFs. Moreover, we present the Hierarchical Semantics Encoding to ground the fine-grained semantic features to 3D Gaussians and further distilled to the scene representation vectors. We conduct extensive experiments on two RL platforms including Maniskill2 and Robomimic across 10 different tasks. The results show that our method outperforms the other 5 baselines by a large margin. We achieve the best success rates on 8 tasks and the second-best on the other two tasks. 

\end{abstract}
\section{Introduction}
\label{sec: intro}
Vision-based reinforcement learning (RL) is receiving increasing attention because vision is a very intuitive way for humans to perceive the world. The typical way for vision-based RL is that an encoder converts the high-dimensional images into a compact, low-dimensional latent state representation. Subsequently, the RL algorithm, such as the policy network, utilizes this latent vector as the input for the state. Hence, the quality of the compact scene representation directly affects the sampling efficiency of vision-based RL. To achieve this, previous approaches usually pre-train the perception encoder on different tasks such as image reconstruction~\cite{islam2022discrete}, contrastive learning~\cite{stooke2021decoupling,liu2021behavior}, and reconstructing task-specific information~\cite{yamada2022task, yang2021representation}. Another alternative way is to add an auxiliary branch for data augmentation~\cite{srinivas2020curl} or other unsupervised tasks~\cite{kinose2023multi, manuelli2020keypoints} to enhance the RL learning procedure. 

Although these approaches positively influence the latent space for improving the data efficiency of downstream RL tasks, they are still constrained by perceiving realistic 3D structures because the 3D geometry of the real world is essential for many RL tasks. Recently, Neural Radiance Fields (NeRFs)~\cite{mildenhall2021nerf} have shown great success in learning 3D scene representations, which implicitly represent scenes within a neural network and can be accessed by arbitrary camera views to render novel views. For the first time,~\cite{driess2022reinforcement} and~\cite{li20223d} use NeRFs to learn the 3D-consistent implicit representation of the environment for robot manipulation. They map multiview images into a latent vector via an autoencoder and adopt conditional NeRF to render back to RGB images for supervision. Furthermore,~\cite{shim2023snerl} proposes to use NeRF to learn both 3D-aware semantic and geometric representations for RL agents by predicting 3D semantic fields with labels. 

However, these NeRF-based representations face issues with data inefficiency and slow training because the volumetric rendering requires dense sampling in the 3D space, and most sampling points are located in empty areas. Moreover, they also do not effectively utilize 3D geometric priors encoded in the RGBD observations. 
The uniform and dense 3D sampling is not geometrical-aware, which does not consider the specific geometric shapes of the scene, further damaging the perception of the scene representation vector to the 3D structural information. 
Besides, learning the semantic information over both free and unoccupied space is generally inefficient, as many human and robotics tasks only require semantic knowledge of occupied space.

3D Gaussian Splatting (3DGS) has brought more attention to the community to overcome the NeRFs' shortcomings. It combines the benefits of explicit and implicit representations. In 3DGS, scenes are represented as point clouds with learnable geometry parameters. One can render novel views by efficient Gaussian rasterization instead of volumetric rendering. Very recently, ~\cite{wang2024reinforcement} and~\cite{lu2024manigaussian} have been the first attempts to integrate 3DGS into RL. However, they either regard 3D Gaussians as scene observations by converting points to Gaussians or directly build RL environments by 3DGS. Both ways demand substantial computational resources to synchronously update the corresponding Gaussian scene representations at each environment state update when training RL policies. Currently, no existing approach exploits the differentiable 3DGS framework for 3D-aware representation learning from only vision input. 

This work proposes the first approach to leverage 3DGS to learn an improved image encoder that maps multiview images into a 3D-aware and semantically-enhanced latent space. First, we assume each Gaussian should carry its semantic feature to reflect (1) which object it belongs to and (2) which part of the object it belongs to. Fine-grained features are beneficial to recognize distinct parts of objects. Hence we beforehand build a hierarchical semantic encoding scheme to guide the training of the GS framework. More importantly, 3DGS represents scenes as points and requires per-scene optimization. The question is how can one render the scene from the latent vector by using the 3DGS technique. To leverage the efficient 3DGS for representation learning, we propose the Query-based Generalizable Feature Splatting (QGFS) that can render the scene from a global latent code that describes the environment without scene-by-scene optimization. 
We summarize our main contributions as follows:
\begin{itemize}
    \item To our best knowledge, We exploit the 3D Gaussian Splatting framework to learn semantic- and geometric-aware scene representation for vision-based RL for the first time. 
    \item We propose the Hierarchical Semantics Encoding (HSE) to guide scene representation learning by Gaussian language fields. Furthermore, we propose the Query-based Generalizable Feature Splatting (QGFS) to render scenes from a single latent code that represents the scene, which enables exploitations of the efficient 3DGS for representation learning. 
    \item Extensive experiments on Maniskill2 and Robomimic platforms show our superior performance over all baselines. We achieved the best results in eight of the ten tasks on both platforms, leaving two with second place.
\end{itemize}
\section{Related Work}
\label{Relate}
\subsection{3D Scene Representations}
There are two primary methods of representing 3D scenes: explicit and implicit. Traditional methods based on point cloud \cite{achlioptasLearningRepresentationsGenerative2018, liuNeuralRenderingReenactment2019}, voxel \cite{lombardiNeuralVolumesLearning2019, sitzmannDeepvoxelsLearningPersistent2019}, and mesh \cite{ thiesDeferredNeuralRendering2019, liuGeneralDifferentiableMesh2020} typically focus on directly optimizing explicit geometric representations. However, due to limitations in resolution and memory, these methods struggle to describe local geometric details efficiently. On the other hand, some methods \cite{mildenhall2021nerf, xuDisnDeepImplicit2019, tancikFourierFeaturesLet2020,jiangLocalImplicitGrid2020}  that implicitly encode scenes through neural networks have received considerable attention. The most notable method, Neural Radiance Field (NeRF) \cite{mildenhall2021nerf}, has been employed across various purposes \cite{yan2023nerf, cao2023hexplane, wu20234d,yuanNeRFeditingGeometryEditing2022, kuang2023palettenerf, yang2022neumesh,mi_switch-nerf_2023,xiangli_bungeenerf_2023,tancik_block-nerf_2022}. 
Recently, \cite{kerbl20233d} proposed 3D Gaussian Splatting (3DGS) which is considered the next generation to represent 3D scenes as point sets with learnable local geometry parameters. 3DGS spurs the development of many fields \cite{chen2024survey}.


\subsection{Representing 3D Scene as Latent Vector}
Earlier methods use a standard convolutional autoencoder to learn 3D-aware representations under given camera poses. \cite{tatarchenko2016multi, worrall2017interpretable, eslami2018neural}. 
NeRF \cite{driess2022reinforcement} is a novel implicit representation of 3D scenes that has garnered significant attention. Based on NeRF, some methods \cite{yu2021pixelnerf,chen2021mvsnerf,wang2021ibrnet} utilize latent-conditioned NeRF to represent the scene as vectors, aiming to achieve generalization and improve the quality of view synthesis. Other studies~\cite{driess2022reinforcement,shim2023snerl,li20223d} extend latent-conditioned NeRF to RL manipulation tasks.~\cite{shim2023snerl} additionally integrates semantic information to the NeRF latent space. 


\subsection{Learnable State Representations for Reinforcement Learning.}
Visual-based reinforcement learning maps images to low-dimensional latent embedding via encoders. Previous methods trained the perception encoder by using various auxiliary objective functions, such as image reconstruction~\cite{islam2022discrete}, contrastive learning~\cite{stooke2021decoupling,liu2021behavior}, and reconstructing task-specific information~\cite{yamada2022task, yang2021representation}, data augmentation~\cite{srinivas2020curl} or other unsupervised tasks~\cite{kinose2023multi, manuelli2020keypoints}. However, these methods struggle to perceive and understand 3D-structural information. Recent approaches directly combine RL and 3D representations to be aware of 3D environments.
NeRF-RL~\cite{driess2022reinforcement} and~\cite{li20223d} first leverage NeRF to train encoders to obtain state representations for RL algorithms.
SNeRL \cite{shim2023snerl} generate semantic-aware representations by learning RGB fields, feature fields, and semantic fields.
The above NeRF-based representation considers view-independent 3D structure information while suffering from inherent disadvantages of the NeRF pipeline.

Another two works are to some extent relevant to this paper. They also incorporate 3DGS into RL. 
GSRL \cite{wang2024reinforcement} proposes a generalizable Gaussian splatting framework for robotic manipulation tasks that can represent local details well without the mask.
ManiGaussian~\cite{lu2024manigaussian} proposes to use dynamic 3DGS to build a digital twin of the per-scene representation of the RL environment.
However, both approaches consider 3DGS as either the observations of the environment or the auxiliary modeling of the scene, which require significant computations. They do not learn a latent representation of the scene with the assistance of the efficient 3DGS framework. 
On the contrary, we adapt the 3DGS to learn the scene representation with hierarchical semantics information. 
\section{Background}
\label{Background}
\subsection{Reinforcement Learning}
We formulate reinforcement learning policy training as an infinite-horizon discrete-time Markov Decision Process (MDP). MDP is defined as a tuple $(\mathcal{S},\mathcal{A},\mathcal{R},p,\gamma)$, where $\mathcal{S}$ is the state space, which includes the RGB and depth images of the whole scenario. $\mathcal{A}$ is the action space for robot arm manipulation, $\mathcal{R}$ is the reward function, $p(s_{t+1}|s_t, a_t)$ represents the probabilities of the transition to the next state for each states-action pair at $t$, $\gamma \in [0,1]$ is the discount factor of reward. At each time step $t$, the policy observes the state $s_t \in \mathcal{S}$ from the environment. The policy executes the action $a_t \in \mathcal{A}$ which is sampled from policy $\pi(a_t|s_t)$. Subsequently, the state transitions from $s_t$ to $s_{t+1}$ by $s_{t+1} \sim p(s_{t+1}|s_t,a_t)$. And the policy obtains a reward value at each time step $r_t=\mathcal{R}(s_t,a_t)$. For maximizing the return discounted reward, the objective is to optimize the parameters of the policy $\theta$: 
\begin{equation}
    \textnormal{arg}\max_{\theta} \mathbb{E}_{(s_t,a_t) \sim p_\theta(s_t,a_t)} \left[ \sum_{t=0}^{T-1} \gamma^t r_t\right]
\end{equation}
where T denotes the time horizon of MDP.

\subsection{3D Gaussian Splatting}

3DGS explicitly models 3D static scenes as 3D Gaussian primitives, each has an opacity ($\alpha_i$), a mean ($\mu_i \in \mathbb{R}^{3} $), a covariance matrix ($\sum_i \in \mathbb{R}^{3\times3} $), spherical harmonics coefficients ($\textbf{SH}_i$) (or a RGB color $c$) and an optional feature vector ($f_i \in \mathbb{R}^{3} $). Among them, to facilitate optimization by gradient descent, $\sum_i$ can be decomposed into a scaling matrix ($S \in \mathbb{R}^{3\times3}$), described by a scale vector ($\mathbf{s}$) and a rotation matrix ($R \in \mathbb{R}^{3\times3}$) described by a quaternion ($\mathbf{q}$):
\begin{equation}
    \Sigma=RSS^{T}R^{T}
\end{equation}
Gaussian rasterization accesses these primitives to render novel views and produce corresponding feature maps.
Given the cameras' parameters and pose matrix, the projection of 3D Gaussians to 2D image plane can be characterized by the view transform matrix ($W \in \mathbb{R}^{3}$) and Jacobian of the affine approximation of the projective transformation ($J \in \mathbb{R}^{3}$), as in:
\begin{equation}
    \Sigma^{'}=JW\Sigma W^{T}J^{T},
\end{equation}
where $\Sigma^{'}$ is the covariance matrix in 2D space. 
By computing the $\alpha$-blend of $\mathcal{N}$ Gaussian points that overlap a pixel projected onto the 2D plane, the final color $C$ and feature $F$ of each pixel is determined:
\begin{equation}
    C = \sum_{i\in \mathcal{N}}c_i\alpha_i\prod_{j=1}^{i-1}(1-\alpha_j),
    F = \sum_{i\in \mathcal{N}}f_i\alpha_i\prod_{j=1}^{i-1}(1-\alpha_j)
\end{equation}
where $c_{i}$ and $f_{i}$ denotes the color and the feature of Gaussian point $i$ ,and $\alpha_{i}$ denotes the soft occupation Gaussian point $i$ at 2D space, which can be calculated by $\alpha_i(x) = o_iexp(-\frac{1}{2}(x-\mu_i)^T{\Sigma}_{i}^{T}(x-\mu_i))$. It is worth noting that ${\Sigma}_{i}$ and $\mu_i$ only represent the 2D-projected version when calculating $\alpha_{i}$. 
\begin{figure*}[t]
    \centering
    \includegraphics[width=\textwidth]{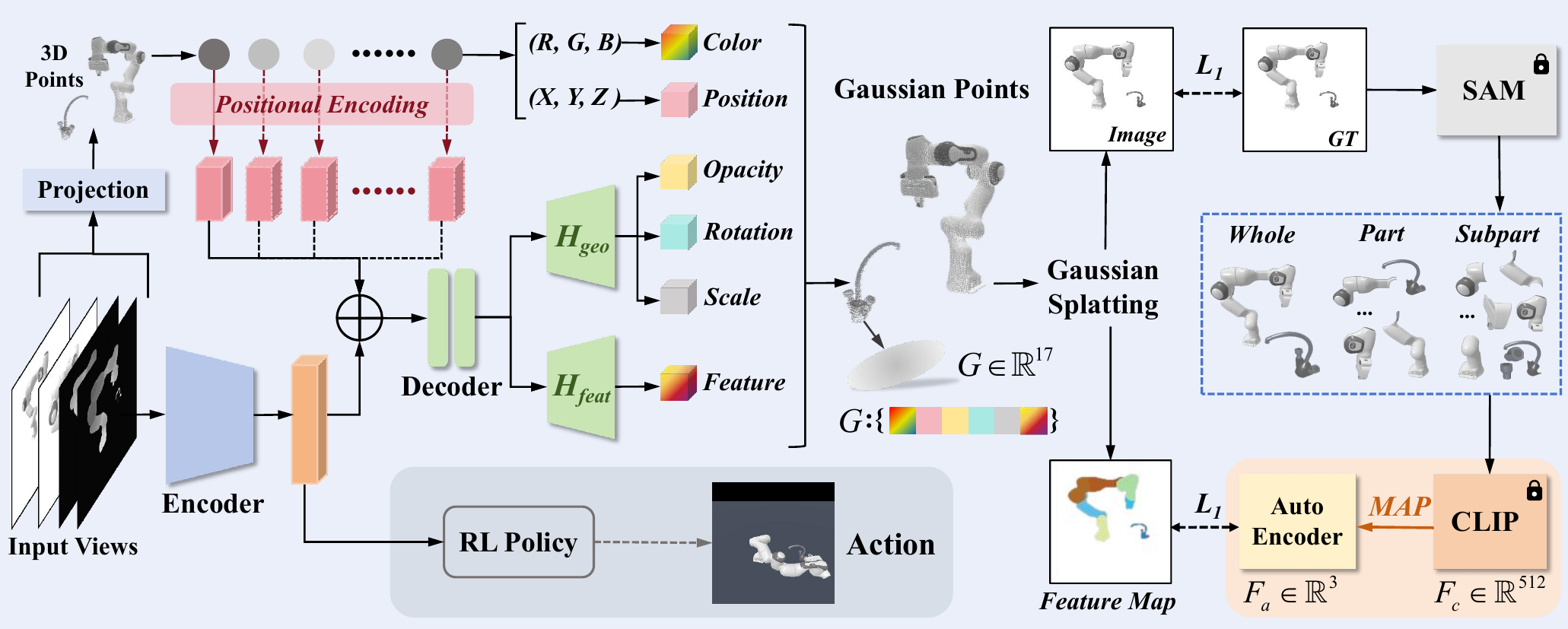}
    \caption{The main pipeline of the proposed approach. The left panel is the Query-based Generalizable Gaussian Splatting, which uses points to query the scene vector to obtain geometry parameters and features and then render novel views via Gaussian Splatting. The right panel is the Hierarchical Semantics Encoding that hierarchically grounds part-level semantic features into the Gaussian field to distill the scene vector.}
    \label{fig: main pipeline}
    \vspace{-0.5cm}
\end{figure*}

\section{Method}
\label{Method}
In this section, we demonstrate the details of the proposed approach which consists of two main blocks, i.e. the Hierarchical Semantic Encoding (HSE) and the Query-based Generalizable Feature Splatting (QGFS). Owing to 3DGS and the hierarchical semantics, our method can extract the scene state vector more efficiently with better semantic and geometric awareness compared to the previous NeRF-based methods, facilitating downstream RL training. The main pipeline of our method can be seen in Fig.~\ref{fig: main pipeline}.
\subsection{Hierarchical Semantics Encoding}
This section introduces the Hierarchical Semantics Encoding (HSE) to ground hierarchical semantics to 3DGS. The previous study~\cite{shim2023snerl} demonstrates that integrating semantic information into latent representation can improve the performance of RL tasks. Further, we note that encoding more fine-grained semantic information should be more beneficial for RL training. For example, when asked to pick up a hammer, a human focuses on the semantic meaning of the handle of the hammer and rarely grabs the hammerhead. Therefore, hierarchical semantic information in the latent vector is beneficial for understanding the scene. Based on this, we ground the semantics to Gaussians hierarchically by the proposed HSE. 

Given an RGB image $I \in R^{H, W, 3}$, we adopt CLIP~\cite{radford2021learning}, the most popular vision-language embedding model, to transform each image into the semantic feature space. It is noted that one CLIP embedding corresponds to an image. Typically, the feature is expanded on the image plane using the object's mask to obtain pixel-level features. As we discussed before, we want to obtain part-level fine-grained language features. Similar to ~\cite{qin2023langsplat, qiu2024feature}, we use the Segment Anything Model (SAM)~\cite{kirillov2023segment} to generate a set of part-level masks for each object and use CLIP to the masked image to produce part-level features. In practice, an independent pixel can belong to multiple part levels. For example, a pixel can belong to the entire cabinet, meanwhile it can also simultaneously belong to the cabinet's handle. Thus we aggregate different levels of part features for each pixel to obtain their final semantic features. In detail, the Masked Average Pooling is adopted to aggregate features. 
\begin{equation}
    \text{MAP}(\mathbf{M}, I)=\frac{\Sigma_{i \in I}\mathbf{M}(i)\cdot \frac{\mathbf{MC}(I|\mathbf{M})}{||\mathbf{MC}(I|\mathbf{M})||}}{\Sigma_{i \in I}\mathbf{M}(i)}
    \label{eq: masked avg pooling}
\end{equation}
where $i$ refers to a certain pixel location. $\mathbf{MC}$ denotes the masked CLIP operation and M is the mask. The output of the MAP operation is assigned to every pixel located within the part segmentation. When a pixel is associated with multiple parts, the pixel feature is derived by averaging the features of all pertinent parts. This provides a part-enhanced CLIP feature map. 

However, CLIP encodes a large number of objects during training. In our specific RL tasks, only a few types of objects would appear, such as robot arms, faucets, cabinets, etc. Hence its original feature space is very redundant for the objects in our RL tasks with common values ranging between $\mathbf{F} \in \mathbb{R}^{H \times W \times 512}$ (for the CLIP-ViT model) and $\mathbf{F} \in \mathbb{R}^{H\times W\times 1024}$ (for the CLIP-ResNet model). In our experiments, we constantly select the CLIP-ViT as our extractor. On account of this, we reduce the dimension of the semantic embedding by utilizing an autoencoder, consisting of an encoder $\mathcal{G}_\phi$ which maps semantic embedding of dimension 512 to the compact latent space $l$, and a decoder $\mathcal{D}_\theta$ that projects the compact latent map back to the original CLIP space. A simple MLP is used as the autoencoder. In this work, we set $l$ as 3 drawing on the reference~\cite{shorinwa2024textbf}. Another by-product of this method is that it maintains the real-time rendering speed of 3DGS. We train the autoencoder with the following loss function:
\begin{equation}
    L_{ae}=\sum_{i \in I}||\mathcal{D}_\theta(\mathcal{G}_\phi(\mathcal{A}(I)))-\mathcal{A}(I)||_2^2
    \label{eq: autoencoder loss}
\end{equation}
where $\mathcal{A}=\sum_m^MMAP(M,I) \cdot MC(I|M)$ denotes the operations to map image $I$ to the compact semantic feature space. After training the autoencoder, we obtain the mapping function from the groundtruth image to the low-dimensional compact latent space, which can be used to ground the hierarchical semantics to 3D Gaussians and will be utilized in the subsequent latent representation learning phase. 
\subsection{Query-based Generalizable Feature Splatting }
In this section, we introduce the QGFS which aims to bridge the 3DGS with scene representation learning. Typical 3DGS has to be optimized in a per-scene manner and represent the scene as points. However, we want to render the scene from a single latent vector that describes it. The QGFS is proposed to render the scene from a vector by using the 3DGS technique in a generalizable way. The main process of QGFS refers to the left part of Fig.~\ref{fig: main pipeline}. 

Similar to previous methods ~\cite{shim2023snerl, driess2022reinforcement}, we adopt the multiview encoder to map multiview observations into a single latent vector $z$ for RL tasks.  
\begin{equation}
    \mathbf{z} = \Omega(o^{1:V}, K^{1:V})
    \label{eq: multi-view encoder}
\end{equation}
The encoder $\Omega$ takes the observed RGBD images $o^i \in \mathbb{R}^{H \times W \times 4}$ and their corresponding camera projection matrix $K^i \in \mathbb{R}^{3 \times 4}$ obtained from $V$ different camera views and outputs the latent code. The detailed architecture of $\Omega$ can be viewed in the Appendix. 

After obtaining the latent code, previous NeRF-based methods conduct volumetric rendering within a NeRF conditioned on $z$. They sample multiple points along rays and predict their color and opacity by an MLP conditioned on $z$, then perform color blending for each ray to obtain their final colors. However, as we discuss in Sec.~\ref{sec: intro}, most sampling concentrates on unoccupied regions, which causes data inefficiency and disturbed geometrical awareness. Therefore, in our method, we reutilize the geometrical priors encoded in the depth observations. We reproject the RGBD images into 3D space by camera projection matrices. 
\begin{equation}
    \mathbf{x} = P \cdot K^{-1} \cdot (u,v,\mathcal{D}(u,v))
    \label{eq. unprojection depth}
\end{equation}
where $K$ and $P$ refer to the intrinsic and pose matrix respectively. $\mathcal{D}(u,v)$ denotes the value at $(u,v)$ on the depth map. 
Then these 3D points are considered keys for querying the global latent vector. As stated in Eq.~\ref{eq: query decoder}, we feed each point coordinate to a positional encoding layer ($\gamma(\mathbf{x})$) and concatenate it with the scene representation vector $z$. Then the concatenation is fed to a lightweight query decoder ($Q_d$) to produce a coordinate-specific feature vector $f_x^{local}$. 
\begin{equation}
    f_x^{local} = Q_d(\gamma(\mathbf{x}) \oplus \mathbf{z})
    \label{eq: query decoder}
\end{equation}
By doing this, each query occurs at the geometry surface of objects, which is efficient and geometry-aware. 
Next, the queried local feature vector is transformed to the semantic features, and Gaussian parameters including opacity, rotation, and scale factors at the position via different prediction heads. 
\begin{equation}
\begin{aligned}
\textbf{q} = norm(\mathcal{H}_r(W_r f_x^{local}+b_r)),\\
\textbf{s} = exp(\mathcal{H}_s(W_s  f_x^{local}+b_s)), \\
o = Sig(\mathcal{H}_o(W_o  f_x^{local}+b_o)),\\
f_x = Tanh(\mathcal{H}_{feat}(W_f f_x^{local}+b_f)) 
\end{aligned}
\label{sec. regression gs param}
\end{equation}
These $W$s and $b$s are the parameters of linear transformation to project $f_x^{local}$ to the Gaussian geometry spaces. 
Additionally, they have different activation functions after the prediction heads to ensure their value range. $norm$ denotes the normalization function along the last dimension. $exp$ is the exponential function. $Sig$ refers to the Sigmoid activation and $Tanh$ denotes the tanh activation. 

After we obtain the four parameters, we keep the original color and coordinate from RGBD projection. Then we can render these queried Gaussian points via rasterization to produce an image and a language feature map given another camera viewpoint that is different from the original input views. The produced two maps are used to compute losses to the groundtruth. 
\begin{equation}
\begin{aligned}
    L_{gs} = \lambda ||I_v - R(\mathbf{x}, c, \mathbf{q}, \mathbf{s}, o|v)||_1^1 + \\
    (1-\lambda)SSIM(I_v, R(\mathbf{x}, c, \mathbf{q}, \mathbf{s}, o|v))
    \label{eq: Gaussian loss}
\end{aligned}
\end{equation}
where $R$ refers to the 3DGS rendering function to generate maps given viewpoint $v$. The $L_{gs}$ loss is the combination of the L1-photometric loss and the structural similarity index measure (SSIM) loss~\cite{wang2004image}. The other loss supervises the results in the feature space.
\begin{equation}
\begin{aligned}
    L_{feat} = \eta||\mathcal{G}_\phi(\mathcal{A}(I_v))-R(\mathbf{x}, \mathbf{q}, \mathbf{s}, f_i, o|v)||_1^1 + \\
    (1-\eta)(1-\psi(\mathcal{G}_\phi(\mathcal{A}(I_v)), R(\mathbf{x}, \mathbf{q}, \mathbf{s}, f_i, o|v))) 
    \label{eq: feature loss}
\end{aligned}
\end{equation}
where $\psi$ denotes the cosine similarity. The feature loss $L_{feat}$ projects the feature map and compares it with the output of the HSE module which we introduce the symbols' meaning in the previous section. The total loss is $L_{total}=\beta_1 L_{gs} + \beta_2 L_{feat}$ where $\lambda_1$ and $\lambda_2$ are weights to balance the two terms. It is noted that in the first 5000 iterations which we called the warm-up stage, we remove the feature loss and only train the geometry head to obtain good geometry reconstruction ability at first. In our experiments, we set $\beta_1=0.4$, $\beta_2=0.6$, $\eta=0.99$, and $\lambda=0.8$.

All in all, in QGFS, we reconstruct the 3DGS by querying the global latent vector with the geometry surface points to obtain their respective Gaussian parameters. In this case, the GS can be built in a generalizable way and the latent vector should retain geometrical and semantic awareness. After training, the encoder $\Omega$ is exploited as a representative feature extractor for any downstream RL task.  
\section{Experiments}
\subsection{Experimental Settings}
In this section, we demonstrate several RL tasks to show the effectiveness of our proposed method compared to other existing approaches. We conduct experiments on two RL platforms, Maniskill2~\cite{gu2022maniskill2} and Robomimic~\cite{mandlekar2021matters} for comprehensive comparisons. In Maniskill2, we select six different manipulation tasks including StackCube, PlugCharger, PickEGAD, PickYCB, OpenDrawer, and TurnFaucet, which we briefly introduced in Fig.~\ref{fig: Maniskill manipulation demonstration}. In Robomimic, we select Lift, Can, Square, and Transport tasks. We follow the default settings implemented in each platform over all experiments and we do not tune the parameters carefully because our goal is to demonstrate our scene representation is general and effective but not to maximize the performance of each RL algorithm. We adopt the DAPG~\cite{rajeswaran2017learning} for all Maniskill2 tasks, a hybrid approach combining the traditional policy gradient technique with learning from demonstrations because Maniskill2 provides some human demonstrations for each task. Moreover, we evaluate all baseline scene representation methods by using BCQ~\cite{fujimoto2019off} and IQL~\cite{kostrikov2021offline} algorithms for the three Lift, Can, and Square tasks of Robomimic, and using the IRIS~\cite{mandlekar2020iris} algorithm for the Transport task.  
For both platforms, in our standard experimental setting, we use 2 cameras to globally observe the scene, which will be used to reconstruct the latent representation of the scene. For fair comparisons, we maintain camera settings constantly the same across all the other baseline experiments. 
\begin{table*}[]
    \centering
    \caption{The success rate of comparisons between ours and 5 baselines on six tasks of the Maniskill2 platform.}
    \vspace{-2mm}
    \setlength\tabcolsep{4.5mm}{
    \begin{tabular}{lcccccc}
        \hline
   Method &StackCube &PickSingleYCB & PickSingleEGAD &PlugCharger &OpenDrawer &TurnFaucet\\
    \hline 
   CNN-AE\cite{finn2016deep}       &0.94±0.03      &0.28±0.14      &0.48±0.08      &0.00±0.00      &0.00±0.00      &0.02±0.02\\
   CURL\cite{srinivas2020curl}       &0.89±0.09      &0.39±0.04      &0.41±0.10      &0.01±0.01      &0.00±0.00      &0.00±0.00\\
   NeRF-RL\cite{driess2022reinforcement}       &0.91±0.06      &\underline{0.44±0.03}      &0.50±0.01      &0.01±0.01      &0.02±0.02      &0.04±0.02\\
   SNeRL\cite{shim2023snerl}       &\underline{0.98±0.02}      &0.42±0.04      &\underline{0.62±0.04}      &\underline{0.02±0.01}      &\underline{0.04±0.00}      &\underline{0.05±0.02}\\
   GSRL\cite{wang2024reinforcement}       &0.95±0.04      &0.42±0.09      &0.61±0.03      &0.01±0.01      &0.02±0.02      &0.04±0.03\\
   Ours       &\textbf{0.99±0.01}      &\textbf{0.46±0.03}      &\textbf{0.70±0.03}      &\textbf{0.03±0.01}      &\textbf{0.07±0.01}      &\textbf{0.08±0.02}\\
    \hline
    \end{tabular}}
    \label{tab: maniskill2}
    \vspace{-2.8mm}
\end{table*}

\begin{table*}
    \centering
    \caption{The success rate of comparisons between ours and 5 baselines on six tasks on the Robomimic platform.}
    \vspace{-2mm}
    \setlength\tabcolsep{6.5mm}{
    \begin{tabular}{lcccccc}   \hline
    &\multicolumn{2}{c}{Lift} &\multicolumn{2}{c}{Can} &Square&Transport\\
   Method &BCQ &IQL &BCQ &IQL &BCQ &IRIS \\ \hline 
   CNN-AE\cite{finn2016deep}      &0.97±0.01  &0.97±0.01  &0.70±0.02      &0.75±0.02   &0.36±0.05  &0.34±0.04\\
   CURL\cite{srinivas2020curl}       &0.97±0.02      &0.98±0.02    &0.69±0.03   &0.74±0.03   &0.41±0.04  &0.26±0.08\\
   NeRF-RL\cite{driess2022reinforcement}       &0.98±0.02   &0.98±0.01    &0.70±0.03   &\textbf{0.77±0.02}   &0.45±0.06 &0.28±0.12\\
   SNeRL\cite{shim2023snerl}      &\textbf{0.99±0.01}   &\textbf{1.00±0.00}    &\underline{0.71±0.03}   &\underline{0.76±0.02}   &\underline{0.49±0.03}   &0.36±0.05\\
   GSRL\cite{wang2024reinforcement}       &\underline{0.98±0.01}   &\underline{0.99±0.01}    &0.70±0.02   &\underline{0.76±0.01} &0.48±0.03   &\underline{0.36±0.02}\\
   Ours       &\textbf{0.99±0.01}   &\underline{0.99±0.01}    &\textbf{0.73±0.02}   &\underline{0.76±0.02}   &\textbf{0.51±0.03}   &\textbf{0.41±0.04}\\
    \hline
    \end{tabular}}
    \label{tab: robomimic}
    \vspace{-4mm}
\end{table*}
\begin{figure}[ht]
    \centering
    \includegraphics[width=0.48\textwidth]{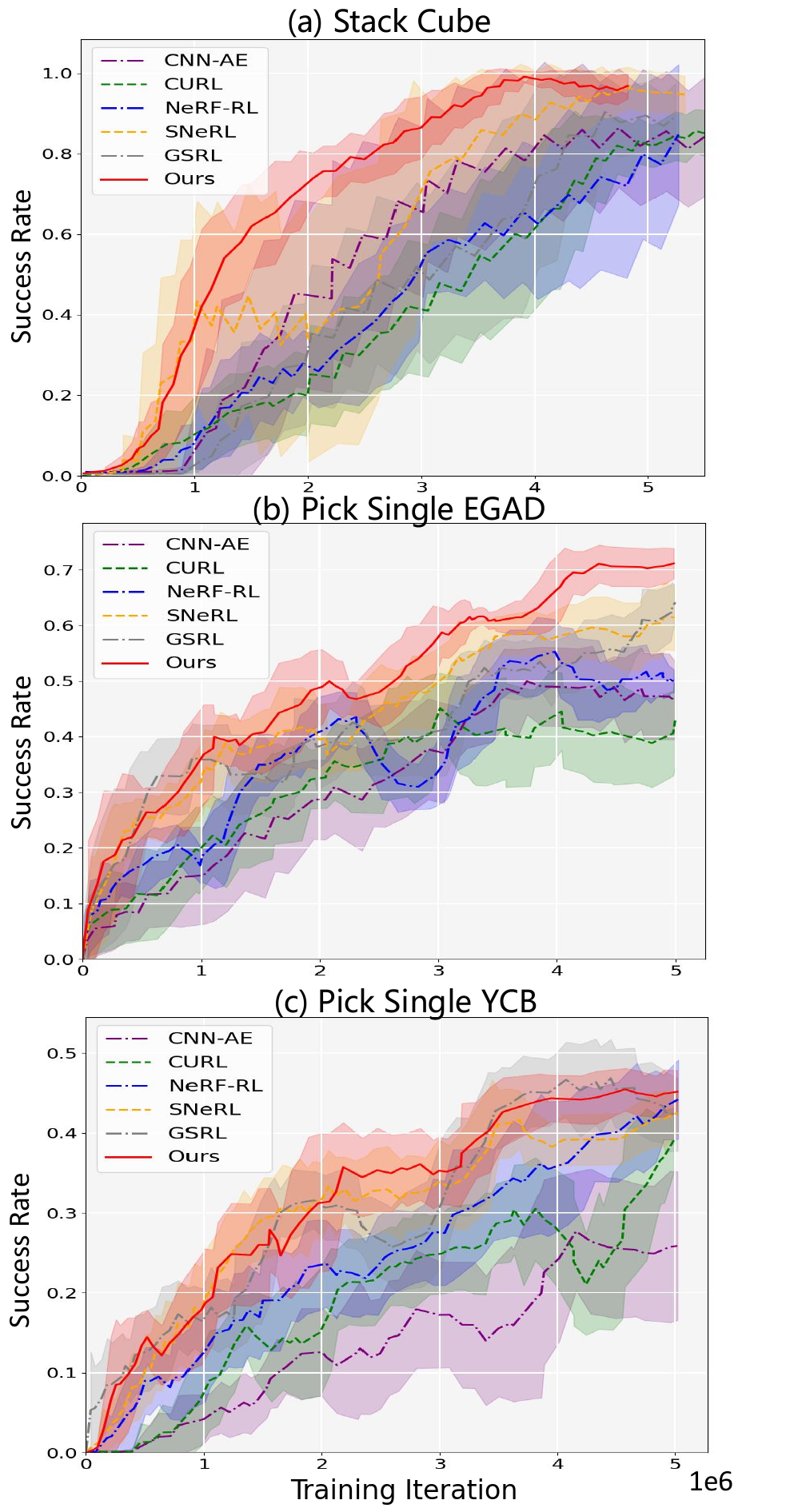}
    \vspace{-0.3cm}
    \caption{Learning Curves Comparisons of ours and the other 5 baselines. }
    \label{fig: training curve on maniskill2}
    \vspace{-0.4cm}
\end{figure}
\begin{figure}[ht]
    \centering
    \includegraphics[width=0.48\textwidth]{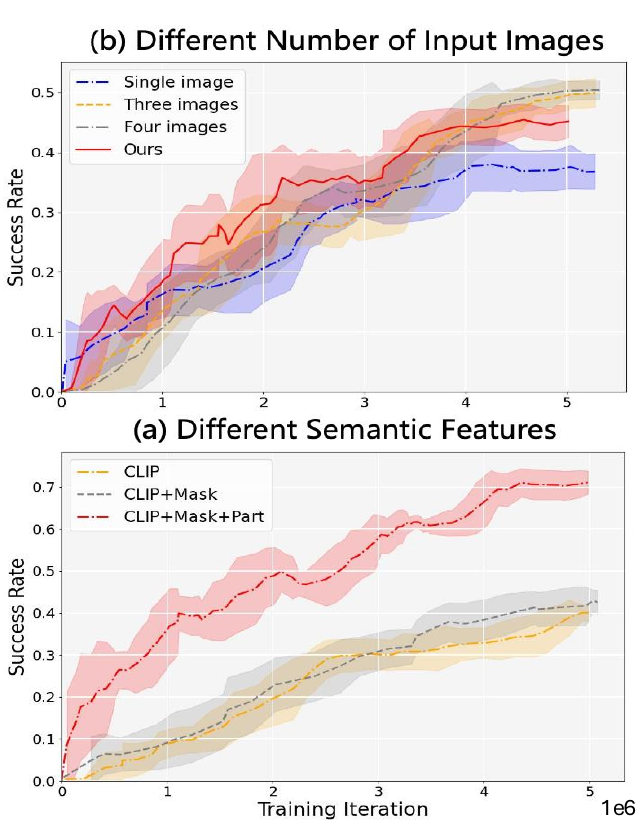}
    \vspace{-0.4cm}
    \caption{The training curves of the ablation studies: (a) the ablation of different settings of semantic encoding, (b) the ablation of the different numbers of input images.}
    \label{fig: Ablation on maniskill2}
    \vspace{-0.4cm}
\end{figure}
\noindent \textbf{Baselines.} We compare our method with other alternative ways of training the encoder for RL, which we briefly described below. CURL~\cite{srinivas2020curl} adopts an auxiliary contrastive loss to train RL, ensuring that the embeddings for data-augmented versions of observations match. Similar to~\cite{shim2023snerl, driess2022reinforcement}, we use its multiview adaptation for better capacity. CNN-AE~\cite{finn2016deep} directly uses the convolutional network as an autoencoder to reconstruct images in the pertaining phase. NeRF-RL~\cite{driess2022reinforcement} pre-trains the encoder with the help of conditional NeRF to obtain the 3D-aware latent vector. SNeRL~\cite{shim2023snerl} adopts additional semantic distillation based on the NeRF-RL by the NeRF-style semantic supervision. Additionally, we compare our method with another recent 3DGS-based RL training scheme GSRL~\cite{wang2024reinforcement}. It directly converts RGBD observations into 3D Gaussians to explicitly represent the scene and uses a point-based encoder to encode the 3D Gaussians before input to the RL policy network. It requires updating the Gaussian representation after each time of interaction with environments, which is computationally expensive. It does not consider using a geometrical latent code and semantic enhancement. 
\subsection{Implementation details}
We pre-train the $\Omega$ in Eq.~\ref{eq: multi-view encoder} by using offline RGBD images. In detail, we collect 12 multiview images for each episode and treat every three spatially adjacent images as a group. We concatenate the first two RGBD images and input them into the encoder to produce the scene representation vector. Then we render the images and the proposed HSE feature maps at the last camera viewpoint from the representation vector by using the presented QGFS to compute losses with the groudturths of the last image.
Beforehand, we pre-trained the Autoencoder with 200 epochs at a learning rate of $5e^{-4} $ for compressing hierarchical semantic features to a compact and low-dimensional feature space, which largely reduces the learning burdens. 
Finally, the $\Omega$ is trained with 10 epochs with the learning rate of $5e^{-5}$. 
All models are trained on a single RTX3090 graphics card using the Adam optimizer. When we train the RL policy, we set the learning rate of the pretrained 3D-aware encoder to be a very small value, i.e. 1e-7 to fine-tune it slightly. 

\noindent \textbf{Encoder.}
The encoder is applied to both the left and right view images and depth maps. Images and depth maps from two views are concatenated and fed into a $4\times4$ convolutional layer to obtain higher-dimensional features. Subsequently, these features are further extracted through a Feature Extraction Block followed by two Residual Blocks. Finally, the dense feature map is obtained through the fully connected layer. 
The encoder has a similar architecture to the encoder in ManiSkill2 \cite{gu2022maniskill2}. 
The extracted features are further used to regress the Gaussian parameters and subsequent reinforcement learning tasks.

\noindent \textbf{Positional Encoding.}
Projecting the images and depth maps results in the generation of a colored point cloud. The color and position of the point cloud are also the color map $\chi_c$ and position map $\chi_p$ of the Gaussian points. Simultaneously, as our method is query-based generalizable feature splatting, we utilize positional encoding \cite{mildenhall2021nerf} to transform the original coordinates into high-dimensional vectors, enhancing the transmission of each point's positional information. However, differently from positional encoding of NeRF \cite{mildenhall2021nerf}, we concatenate the results of the positional encoding with the original point position and colors to obtain the final per-point information.

\noindent \textbf{Decoder and Regression Head.}
To provide features used for point queries with more semantic and spatial information, we have designed a lightweight decoder. This decoder simply includes two $1\times1$ convolutions and is used to decode pixel-wise Gaussian features. Then, this feature regresses Gaussian point parameters by three specific Gaussian parameter Regression heads. Specifically, we obtain scaling map $\chi_s$, rotation map $\chi_r$, and opacity map $\chi_\alpha$ through two $3\times3$ convolutions and their corresponding activation functions, respectively.

\subsection{Results Discussion}
Table~\ref{tab: maniskill2} reports the quantitative results, i,e, the mean and variance of success rate, of the six methods on Maniskill2 tasks. Fig.~\ref{fig: training curve on maniskill2} demonstrates the training curves of the six methods on the first three tasks. It can be observed that our approach overall outperforms the other five baseline methods across all tasks thanks to our better geometrical and semantic scene representation. In addition, our method trains more stably because the variance in our success rates is relatively small. It is seen from the training curve that our method converges faster than others because we fully utilize the geometry priors encoded in the depth maps, only sample points to query the scene on the geometry surfaces of the objects. Therefore, our method has higher data efficiency compared to those NeRF-based encoders. In addition, Table~\ref{tab: robomimic} lists the mean and variance of success rates on the tasks of the Robomimic platform. In these six experiments, our method achieved the best performance in at least four of them and secured at least second place in the remaining experiments. 
Interestingly, our advantage over SNeRL is more evident in complex tasks. This is because we leverage the 3DGS framework to learn a more geometrical-aware scene representation while SNeRL relies on the NeRF supervision that uniformly samples points in the whole scene. Furthermore, in tasks of picking up some tools from obstacles including PickSingleYCB or PickSingleEGAD, our model can deliver stable and fast convergence because our scene latent code contains hierarchical semantics that can be used to guide the policy network focus on plausible correlation areas.

\subsection{Ablation Studies}
In this subsection, we ablate some significant designs of the proposed method. To validate the effectiveness of our hierarchical semantics encoding, we set two additional experimental settings when extracting language features. First, we only use CLIP to generate a single latent code for all pixels in this image. Second, we use SAM to produce object-level masks and adopt masked CLIP to produce the latent vector for each object. The pixels belonging to the same object share the same latent feature. Third, we maintain our HSE to simultaneously utilize clip, SAM, and part-level feature aggregation. Notably, the Autoencoder in Eq.~\ref{eq: autoencoder loss} is constantly used for all experiments to keep the rendering speed of 3DGS. Fig.~\ref{fig: Ablation on maniskill2} (a) illustrates the training curves of the three situations. We mark the first case as $clip$ only, the second case as $clip+mask$, and the third case (which is the same as HSE in the main text) as $clip+mask+part$. It can be seen from this figure that the part-level, object-centric semantic features are indeed beneficial for learning better RL policy. 

Furthermore, we evaluate the proposed method with more image observations. In the main experiment, we only adopt two cameras to observe the scenes. In the ablation section, we additionally test our performance when the number of images is 1, 3, and 4. The performance is recorded in Fig.~\ref{fig: Ablation on maniskill2} (b).
A single image of the scene can lead to a decline in algorithm performance. Additionally, there is a trend where performance improves as more images are available. Moreover, the impact of adding more images subsequently is not as significant as the impact of increasing from one to two images.

\section{Conclusion and Limitation}
\noindent \textbf{Conclusion.} To address the drawbacks of NeRF-related scene representation methods, this work presents a novel approach that adopts the 3DGS technique for scene representation learning for the first time. The goal is to learn an efficient and effective latent vector that describes the scene to improve the downstream vision-based RL tasks. In brief, this work proposes two main technical contributions. To ground fine-grained semantic information to the latent scene representation, we adopt the Hierarchical Semantics Encoding to supervise the training of feature splatting in 3DGS. Typical 3DGS aims to reconstruct 3D structures from multiview images and requires per-scene optimization. To adapt 3DGS to scene representation learning, we propose the Query-based Generalizable Feature Splatting that encodes a scene into a latent vector and renders novel image and feature maps from this latent code and depth prior. We conduct experimental evaluations on two RL platforms, Maniskill2 and Robomimic, containing 10 different tasks and make comparisons with 5 different baselines. The results show that our method achieved the best results in eight of the tasks, and at least second place in the remaining two tasks. 

\noindent \textbf{Limitation.} Although satisfactory improvements are achieved, there are still some limitations remaining. We need multiview images to pre-train the semantical and geometrical aware encoder, which might be difficult to collect in some complicated tasks. Second, this method still requires training from scratch for each RL task. In the future, we plan to extend to multi-task generalization or open-vocabulary manipulation by utilizing the rich semantic information.

\bibliographystyle{plain}
\bibliography{main}

\end{document}